 \documentclass[letterpaper, 10 pt, conference]{ieeeconf}
\usepackage{flushend} 

\usepackage{graphics} 
\usepackage{epsfig} 
\usepackage{mathptmx} 
\usepackage{times} 
\usepackage{amsmath} 
\usepackage{amssymb}  

\IEEEoverridecommandlockouts 

\title{\LARGE \bf Energy-efficient tunable-stiffness soft robots \\ using second moment of area actuation}
\author{L.Micklem$^{1}$\thanks{Corresponding author {\tt\small l.micklem@soton.ac.uk}. Manuscript received: February, 24, 2022. $^{1}$Southampton Marine and Maritime Institute, University of Southampton, UK, $^{2}$Alan Turing Institute, London, UK, {$^{3}$Institute of Industrial Science, The University of
Tokyo, Japan}}, G.D.Weymouth$^{1,2}$, B.Thornton$^{1,3}$}

\date{}

\begin{document}

\maketitle

\begin{abstract} 

The optimal stiffness for soft swimming robots depends on swimming speed, which means no single stiffness can maximise efficiency in all swimming conditions. Tunable-stiffness would produce an increased range of high-efficiency swimming speeds for robots with flexible propulsors and enable soft control surfaces for steering underwater vehicles. We propose and demonstrate a method for tunable soft robotic stiffness using inflatable rubber tubes to stiffen a silicone foil through pressure and second moment of area change. We achieved double the effective stiffness of the system for an input pressure change from 0 to 0.8\,bar and 2\,J energy input. We achieved a resonant amplitude gain of 5 to 7 times the input amplitude and tripled the high-gain frequency range compared to a foil with fixed stiffness. These results show that changing second moment of area is an energy effective approach to tunable-stiffness robots.

\end{abstract}

\vspace{5mm}
\section{Introduction}

\begin{figure*}[htbp]
    \centering
    \includegraphics[width=\textwidth]{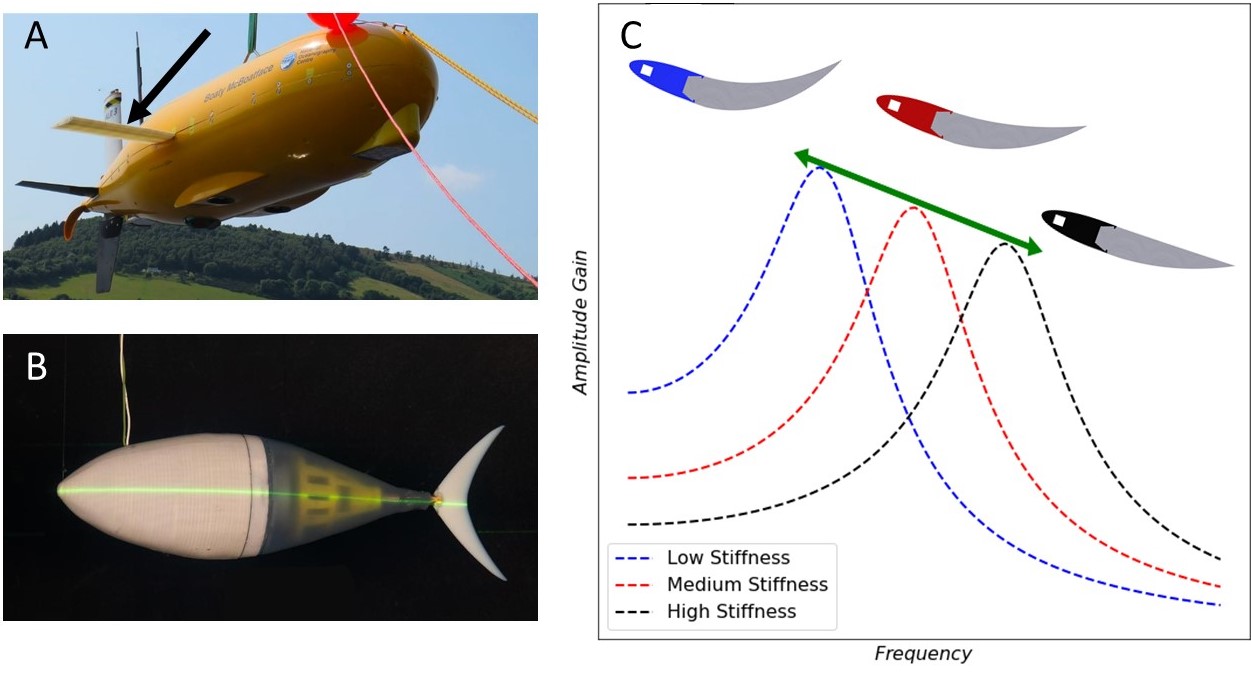}
    \caption{(A) Autosub Long Range, Boaty McBoatface, is an autonomous underwater vehicle with large rigid control surfaces for steering and stabilisation indicated by the black arrow. (B) Tunabot, is a bio-inspired robotic swimmer with dynamic tail actuation using a system of rigid linkages \cite{Zhu2019}. (C) The amplitude gain for three levels of stiffness is plotted against frequency. The peak amplitude of each stiffness is the resonant, and most efficient, frequency for that stiffness. This means being able to tune the stiffness of a system gives a larger operational envelope of efficient frequencies at which to operate.}
    \label{fig:intro_plot}
\end{figure*}

Soft robots are becoming prominent in a variety of fields \cite{Hoang2021,Drotman2021,Umedachi2016,Porez2014,Naclerio2021}. One area of great potential for soft robots is efficient actuation in subsea environments \cite{Li2021,Aracri2021,Youssef2022}. Many pieces of the puzzle to achieve this potential are already in place: flexible strain sensors for deformation feedback have been successfully integrated into flexible systems \cite{Sengupta2020}, open-loop On-Off control has been implemented \cite{Christianson2020,Schegg2022}, and the ability to achieve large amplitude deformations has been demonstrated \cite{Arienti2013}. A key missing piece is the ability to tune the stiffness of soft materials for use as controlled surfaces and actuators, which are needed to steer and propel robots in fluid environments \cite{Al-Rubaiai2021}. Tunable control surfaces would allow truly soft robots to demonstrate and potentially exceed maneuvering capabilities seen in flight-type underwater vehicles such as the Autosub Long Range in Fig.~\ref{fig:intro_plot}(A), and tunable flexible fins would expand the range of efficient swimming speeds for bio-mimetic robots such as Tunabot in Fig.~\ref{fig:intro_plot}(B). 
\bigskip

Flexibility is important for efficient animal swimming, and by extension, swimming robots \cite{Quinn2022}. By being flexible, swimmers are able to make use of resonance. They will deform most easily when actuated at their resonant frequency, and energy that is converted to useful work is maximised. In other words, the amplitude gain is maximised. Some jellyfish save 30\% of their energy costs by actuating at the resonant frequency of their bells and this has inspired the development of high efficiency underwater robots \cite{DEMONT1988,Bujard}.  However, a robot with a fixed stiffness has only one resonant frequency where the amplitude gain and high efficiency operation is maximised. In contrast, a robot with tunable stiffness can maximise gain across a wide range of frequencies, greatly improving its efficient operational envelope, as sketched in Fig.~\ref{fig:intro_plot}(C).

Being flexible also enables the robot to tune its fluid loading; for example, deforming in ways that optimise the lift to drag ratio. This is true in both air and water, as can be seen when leaves deform more and more as wind speed increases to maintain low drag \cite{DeLangre2008}. Swimming animals optimise this hydrodynamic effect by adjusting their swimming amplitude and frequency depending on their speed. We can describe this using the Strouhal number (St). It is an important dimensionless number that defines the ratio of swimming amplitude and frequency to swimming speed St~=~$\frac{Af}{U}$. Most animals swim in a narrow range of St~=~0.25--0.4, but achieve high swimming efficiency over a wide range of swimming speeds \cite{Gemmell2013}. This means that their swimming frequency must increase or decrease with the speed, and must do so without a loss of efficiency. The secret is the tunable stiffness of these animals. Zhong et al. \cite{Zhong2021} have shown experimentally that tuning the stiffness of their tuna-like robot allowed them to double their swimming efficiency at speeds from 0 to 2 body lengths per second. Quinn and Lauder \cite{Quinn2022} explain how fish tune their stiffness through muscle contraction, fin and tail shape alteration, and skin surface changes. They compare this to current swimming robot stiffening methods including structural control \cite{Huh2012}, mechanical control \cite{Li2018}, and intrinsic rigidity control \cite{Behbahani2017}. However, none of these offer a simple mechanical approach to efficiently and continuously adjust the stiffness of the system.

We propose and test a new mechanism that is capable of efficient and continuous stiffness adjustment using soft materials. Our method is based on the inflation of rubber tubes which increases the second moment of area of the tubes thereby geometrically increasing their bending stiffness. When integrated in a soft silicone housing (Fig.~\ref{fig:Foil_CAD}), this allows for large amplitude deformations at a range of frequencies for efficient swimming at a range of swimming speeds. Our aim is to provide a soft method for increasing the operational envelope of efficient frequencies as shown by the green line in Fig.~\ref{fig:intro_plot}.

\vspace{5mm}
\section{Physical models and motivation for second moment of area-based stiffness tuning}

We can model the effect of stiffness by treating soft robotic systems as single degree of freedom mechanisms. In the steady case (such as control surface manipulation, Fig.~\ref{fig:intro_plot}A), the deflection amplitude $A$ is governed by Hook's law $A=F/k$, where $F$ is the applied load and $k$ is the effective stiffness. 

In the case of oscillating deflection (such as propulsive flapping, Fig.~\ref{fig:intro_plot}B), the undamped resonant or natural frequency $\omega _n$ is given by
\begin{equation}\label{eqn:nat_freq}
    \omega _n = \sqrt{\frac{k}{m}}
\end{equation}
where $m$ is the effective oscillating mass. When forced oscillations are applied, the undamped response amplitude gain is given by \begin{equation}\label{eqn:undamped_amp}
    \frac{Ak}{F} = \frac{1}{|1-(\omega/\omega_n)^2|}
\end{equation}
where $F,\ \omega$ are the amplitude and frequency of the applied oscillation. The amplitude gain per unit excitation is the highest when the driving frequency matches the natural frequency, plotted as the dashed lines in Fig.~\ref{fig:intro_plot} (C), but this results in a narrow window of high efficiency operations. By tuning the stiffness $k$, we can adjust the natural frequency and achieve the peak mechanical gain across a wide operational envelope. 

Inflatable supports have been used to stiffen structures from space probes to robotic grippers \cite{Leonard1960,LiMeng2021}. However, their focus is switching between rigid and flexible states. New methods are required for their application to continuous stiffness control. 

\begin{figure}[t!]
    \centering
    \includegraphics[width=8cm]{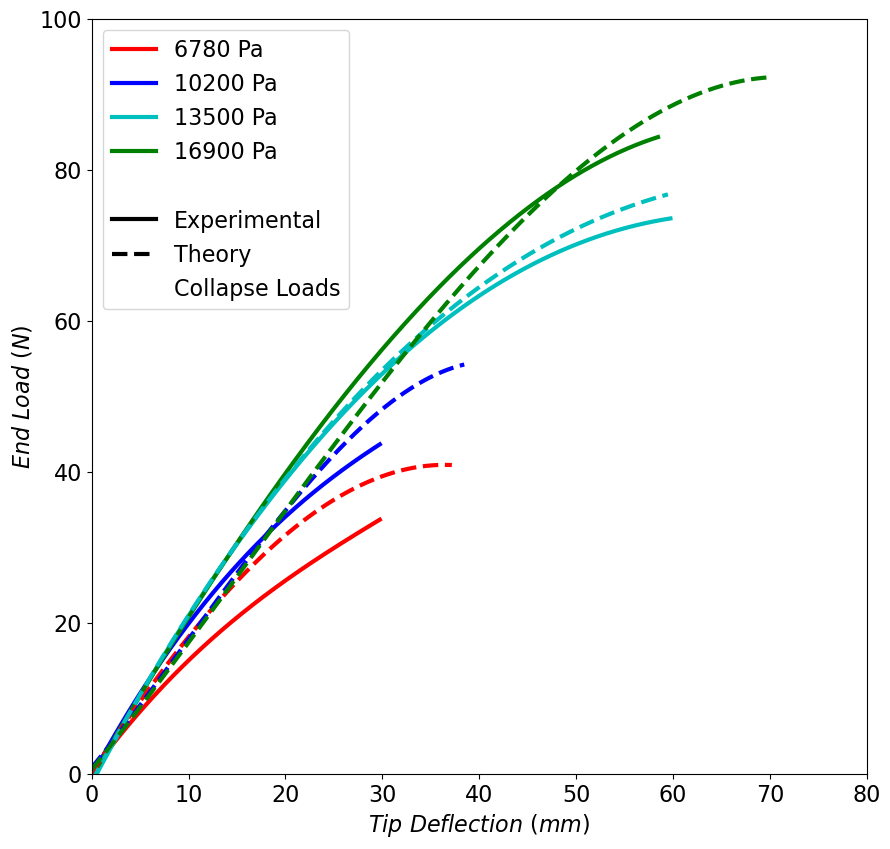}
    \caption{Before the onset of buckling in an inflatable cantilever beam made of in-extensible fabric, there is a geometric stiffness limit. Increasing the pressure in the beam only delays the onset of buckling. Reproduced from \cite{Webber1982}.}
    \label{fig:beam_theory}
\end{figure}

An inflatable cantilever beam made of in-extensible fabric relies on internal pressure to tension the fabric and resist compressive stress on the beam \cite{Comer1963, Webber1982,Okda2019,Zhu2008}. However, the internal pressure doesn't change the geometrically determined stiffness of fabric tubes until near buckling, as shown by the collapse of the deflection vs load behaviour in Fig.~\ref{fig:beam_theory}, \cite{Webber1982}. Pre-buckling, \cite{Leonard1960} shows the tip deflection $A$ of a beam for given end loads $F$ is governed by classical beam theory as

\begin{equation}\label{eqn:tip_deflection}
    A = \frac{FL^3}{3EI}
\end{equation}
where $L$ is the length, $E$ is the material stiffness, and $I$ is the second moment of area which for a hollow tube is $r^3 t$ where $r,t$ are the tube radius and wall thickness. Fig.~\ref{fig:beam_theory} shows this theory trends closely with the experimental deflection until near the buckling load $F_b \approx \left( \frac{\pi p r^3}{2L}\right)$. Therefore, only when operating at extreme pressures near buckling failure would a fabric tube have a controllable deflection, making it both dangerous and expensive to use to tune stiffness.

However, the situation improves dramatically if the inflatable beam is elastic and allowed to expand. In this case, the onset of buckling will be delayed by both the pressure increase and the increase in radius, which will also increase the second moment of area, and therefore the overall beam stiffness $k=F/A \propto EI$ from Eqn.~\ref{eqn:tip_deflection}. Assuming the value of material stiffness and length remain constant the relative change of stiffness of the elastic tubes with pressure is given by

\begin{equation}
    \Delta k/k_0 = 2p+p^2
\end{equation}

\noindent where $p$ is the dimensionless pressure defined as $p=\frac {Pr_0}{Et_0}$ where subscript 0 denotes resting length values of radius and thickness. The conducted experiments aim to establish the effect of this tube stiffening on the overall system.  

\vspace{5mm}
\section{Materials and Methods}

\begin{figure}[t!]
    \centering
    \includegraphics[width=8cm]{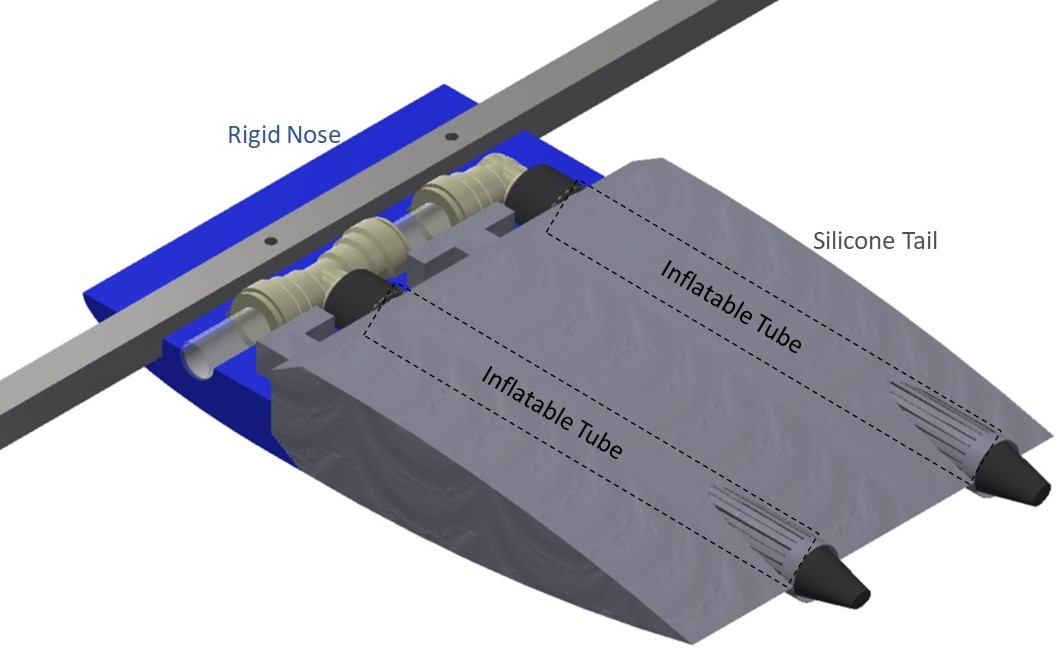}
    \caption{Schematic of the tunable-stiffness foil. The rigid nose attaches to a rotating bar, houses the internal pressure tubing, and clamps the silicone tail. The soft silicone tail has holes to house the inflatable rubber tubes which can expand and contract with pressure.}
    \label{fig:Foil_CAD}
\end{figure}

A tunable-stiffness foil utilising the concept of second-moment-of-area actuation was designed, built, and tested for this study (Fig.~\ref{fig:Foil_CAD} and Table~\ref{table_materials}). The foil is comprised of a rigid nose connected to a soft tail. Embedded within the tail are two inflatable elastic tubes. The tail has a base stiffness provided by the silicone, and the tubes can be pressurised to increase the second moment of area and stiffness. The focus of the study is to demonstrate the relationship between pressure and stiffness, which is independent of the inflation fluid. While we plan to pressurise with water to reduce the buoyancy and hydrostatic pressure effects in the eventual underwater applications, we chose to simplify this initial study by pressuring with and testing in air. One practical issue with using pressure to increase the tube radius is the natural deformation instability of an inflating elastic membrane \cite{Khayat1992,Patil2016}. It is important to prevent these instabilities since a uniform second moment of area increase is needed to provide a uniform stiffness increase. In this work, we achieved uniform inflation by reinforcing the tube walls around the areas of instability with elastic thread, increasing the local hoop strength.

\begin{figure}[t!]
    \centering
    \includegraphics[width = 8.2cm]{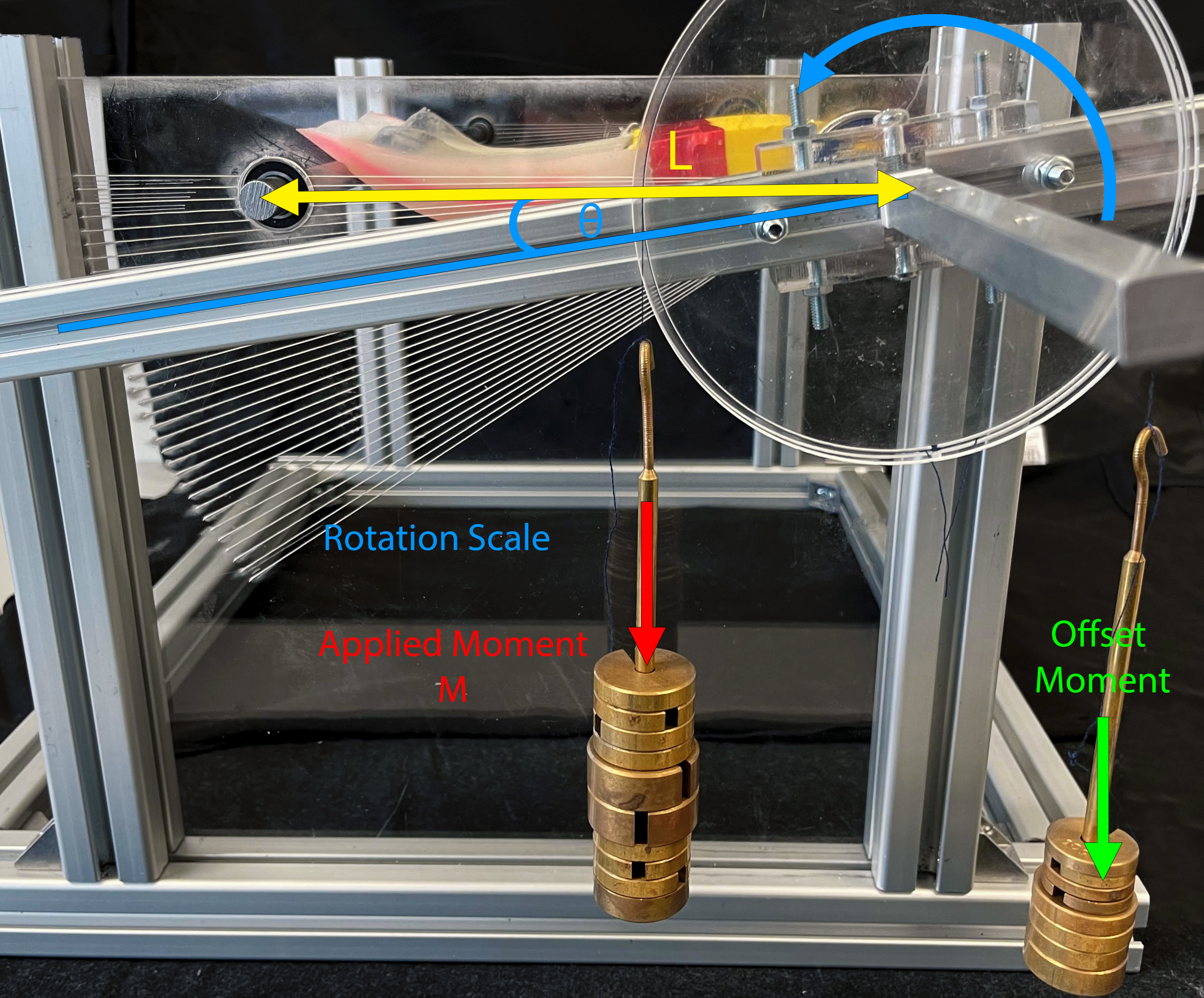}
    \caption{Schematic of the static testing set up. A moment wheel is attached to the rotating square bar. An offset moment is applied to counter the moment caused by the mass of the tunable-stiffness foil. The rear of the foil is resting on a circular bar which is free to rotate. A moment is applied to the square bar and the arm indicates the angle of rotation on the scale.}
    \label{fig:Static_test}
\end{figure}

\begin{table}[ht]
\caption{Experimental Materials}
\label{table_materials}
\begin{center}
\begin{tabular}{c c c }
\hline
 \multicolumn{3}{c}{Stiffness foil}\\
 \hline
 \emph{Part} & \emph{Material} & \emph{Dimensions (mm)}\\
 
Soft Tail & EcoFlex-30 & 120 x 140 x 30\\
Rigid Nose & Polylactic Acid Plastic & 120 x 80 x 30\\
Inflatable Tubes &  Isobutylene Isoprene Rubber & 110 x 15 x 15\\
Square Bar & Aluminium & 700 x 10 x 10\\
\hline

\multicolumn{3}{c}{Dynamic Testing}\\
\hline
Motor & Crouzet brushed DC motor &\\
Motor Driver & Cytron MD10C &\\
Controller & Arduino Uno&\\
Camera & GoPro HERO 7&\\
Resolution & 1920 x 1080&\\
Frame Rate & 240 fps&\\

\end{tabular}
\end{center}
\end{table}
Fig.~\ref{fig:Static_test} shows our experimental setup for static testing of the tunable-stiffness foil for use as a tunable control surface. It is a modification of the classic three-point bending test. The rigid nose section was mounted to a square bar which was free to rotate. The tail rests on a circular bar which is also free to rotate. A moment, $M$, was applied at the axis of rotation which caused the foil to bend through an angle, $\theta$. The angle of rotation was indicated by the arm on the rotation scale. An offset moment was applied to the square bar to zero the rotation induced by the mass of the foil itself. We tested the foil with internal pressures from 0 to 0.8\,bar, in increments of 0.1\,bar. The moment was increased until the foil could no longer support itself and the tail rolled off the circular back bar.

\begin{figure}[t!]
    \centering
    \includegraphics[width=8cm]{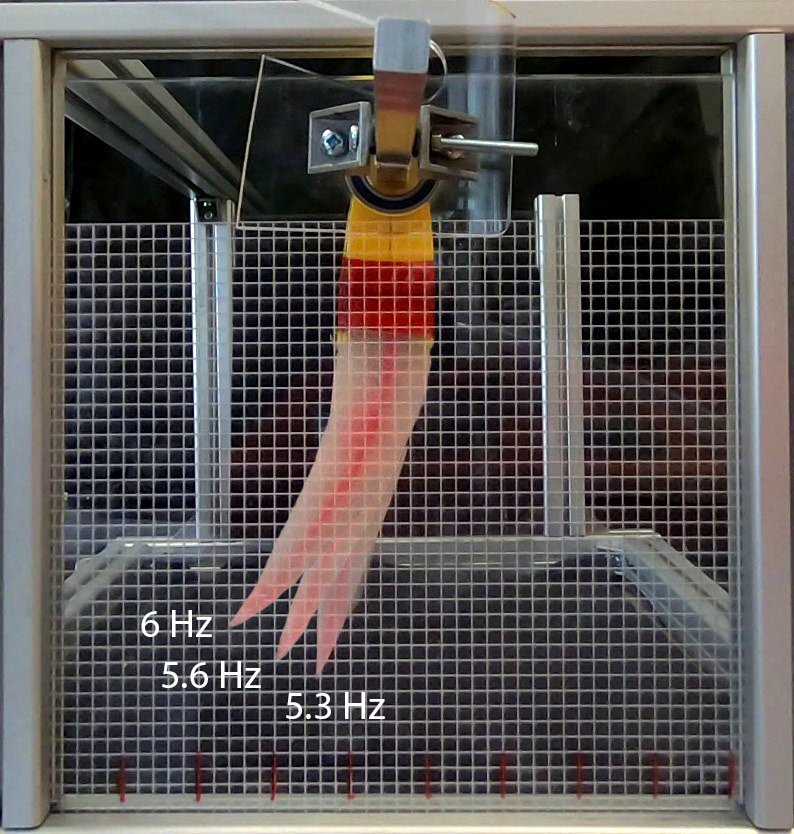}
    \caption{Overlay of different frequency amplitudes during the dynamic testing. The tunable-stiffness foil is driven at a fixed 6\textdegree~peak to peak amplitude. A frequency sweep shows the increase in amplitude as the foil tends towards its natural frequency at a given pressure.}
    \label{fig:Dynamic_Test}
\end{figure}

To test the tunable-stiffness foil dynamically as a tunable flexible fin, we designed a fixed amplitude, frequency sweep experiment (Fig.~\ref{fig:Dynamic_Test}). We oscillated the foil at a fixed 6\textdegree~peak to peak amplitude using a crank shaft mechanism, starting at 1\,Hz, until we reached the maximum frequency achievable in this first mode shape for this experimental set up. We tested the foil at 0, 0.2, 0.5, and 0.8\,bar pressures, using the grid to visually measure the amplitude of deflection. We verified the actuation frequency and amplitude using a video recording of the motion using the camera specified in Table~\ref{table_materials}.

\vspace{5mm}
\section{Results and Discussion}

\begin{figure}[t!]
    \centering
    \includegraphics[width=7.1cm]{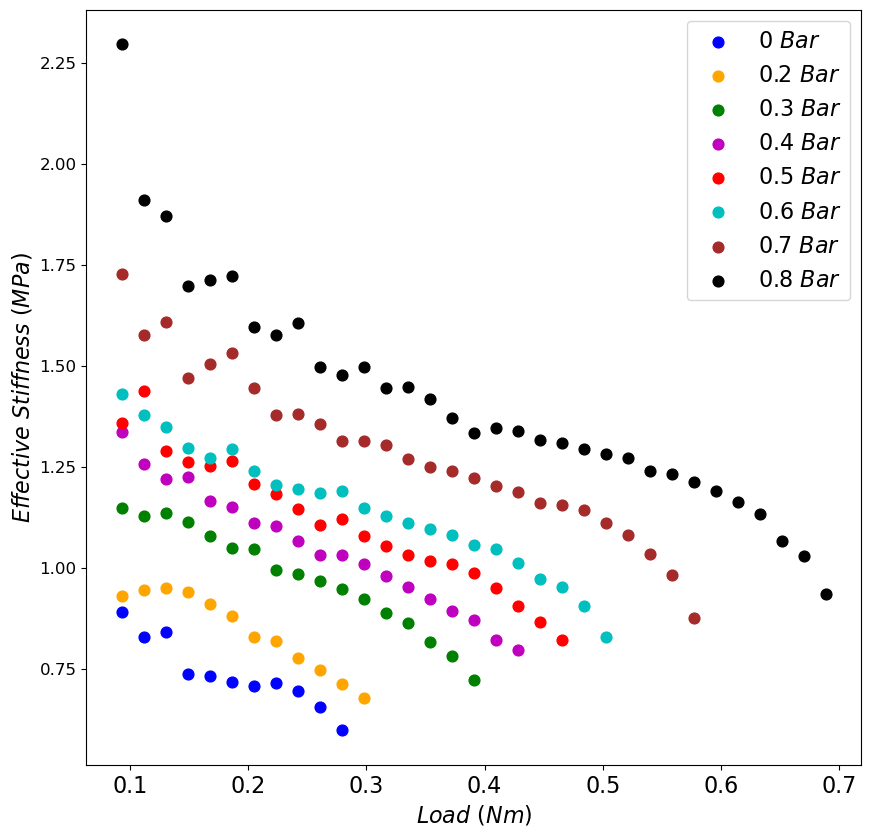}
    \caption{Plot of the effective stiffness of the tunable-stiffness foil against applied moment for 0 to 0.8\,bar pressure. Effective stiffness was calculated from measured deflection using Eqn.~\ref{eqn:effective_stiffness}. Increasing the pressure from 0 to 0.8\,bar sees stiffness increase by a factor of two. By choosing the internal pressure, the stiffness may be tuned to a given value between these limits giving a truly tunable soft control surface.}
    \label{fig:Static_Results}
\end{figure}

\begin{figure}[t!]
    \centering
    \includegraphics[width=7.1cm]{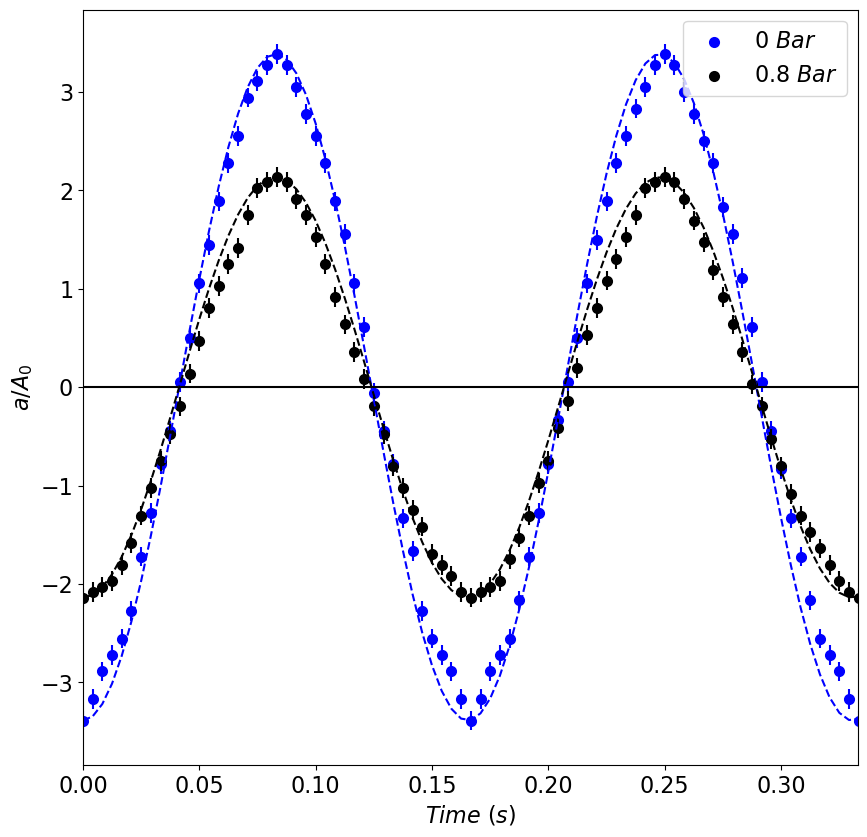}
    \caption{Time history plot of instantaneous normalised tip deflection at 6\,Hz oscillation frequency for 0 and 0.8\,bar pressures. The error bars indicate the accuracy of the measurements allowed by the camera resolution and frame rate (Table.~\ref{table_materials}) when processing the footage.  }
    \label{fig:time_history}
\end{figure}

\begin{figure}[t!]
    \centering
    \includegraphics[width=8.0cm]{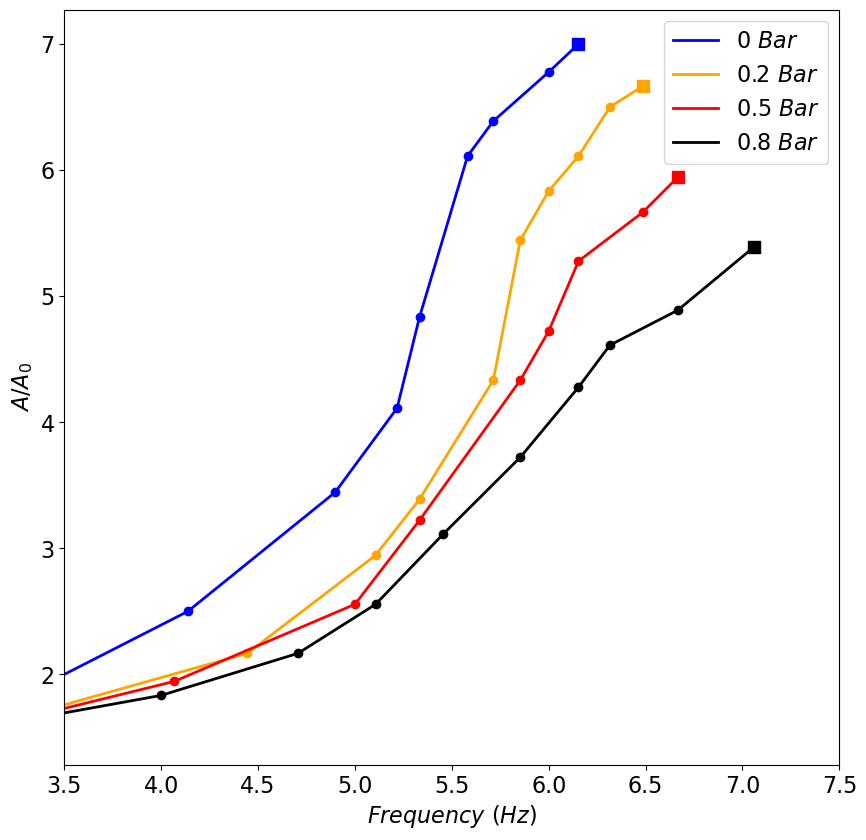}
    \caption{Plot of tail amplitude gain against frequency for 0, 0.2, 0.5, and 0.8\,bar pressures. The square marker indicates the highest amplitude achieved in the first mode shape for this experimental setup. Results show an amplitude gain of 5 to 7 occurring across a range of 1.5\,Hz, three times the frequency range of a single stiffness.}
    \label{fig:Dynamic_Results}
\end{figure}
Fig.~\ref{fig:Static_Results} shows the results of the static stiffness testing. We plot the effective stiffness of the foil using the previously introduced cantilever beam model, Eqn.~\ref{eqn:tip_deflection}. Geometrically, the effective tip load is $F=\frac{M}{L}$ and the tip deflection is $A=L\sin{\theta}$, and so the effective stiffness of the system can be calculated as
\begin{equation}\label{eqn:effective_stiffness}
    E_\text{eff} \equiv \frac {EI}{I_0}= \frac{ML}{3I_0\sin{\theta}}
\end{equation}
where $I_0$ is a nominal averaged second moment of area for the foil when unpressurised.

Fig.~\ref{fig:Static_Results} shows a clear increase in stiffness with each increase in pressure. It is possible to more than double the stiffness of the system by increasing the pressure from 0 to 0.8\,bar. We have also shown that this is a non-binary situation, with each intermediary stiffness also attainable by adjusting the pressure accordingly. This shows the ability of this system to be a tunable soft control surface.  

Fig.~\ref{fig:time_history} shows a time history plot of instantaneous tip deflection, $a$, measured during the dynamic testing. The amplitudes are normalised by, $A_0$, the 1\,Hz peak to peak amplitude at each given pressure. The plot shows the time history for 0 and 0.8\,bar pressures at 6\,Hz oscillation frequency. The motion is reasonably sinusoidal with low cycle-to-cycle variability. The asymmetry in the shape of the 0\,bar peaks can be explained by a slight asymmetry in the manufacture of the tunable-stiffness foil. Increasing the internal pressure to 0.8\,bar reduces the tip deflection amplitude, as expected. 

Fig.~\ref{fig:Dynamic_Results} shows the tail amplitude gain against frequency for 0, 0.2, 0.5, 0.8\,bar pressures. The amplitude gain is the measured tip peak to peak amplitude, $A$, normalised by $A_0$. The square markers indicate the highest amplitude achieved in the first mode shape for this experimental setup. Beyond this point, the system switches to small amplitude second-mode response (a full wave along the foil instead of a half-wave), with no discernible first-mode amplitude. Prior to this mode shift, it is shown that oscillating the system at close to its natural frequency gives an amplitude gain of 5 to 7. We have also shown that by adjusting the internal pressure, and thus the stiffness of our system, we can achieve an amplitude gain of 5 across a range of more than 1.5\,Hz, tripling the high-gain frequency range compared to a foil with a fixed stiffness. Comparing to Fig.~\ref{fig:intro_plot} (C), where Eqn.~\ref{eqn:undamped_amp} is plotted for three values of stiffness, we can see that we have the same increase in the operational envelope of the foil by adjusting the stiffness.    

\vspace{5mm}
\section{Conclusion and Robotic Implications}

We have demonstrated the ability to double the effective stiffness of a soft tunable-stiffness foil by increasing the internal pressure of inflatable tube stiffeners from 0 to 0.8\,bar. The increase in pressure increases the second moment of area of the tubes, which is the key mechanism to efficiently change the system stiffness. Indeed, we achieve a doubling of effective foil stiffness with an energy cost of just over 2\,J due to the increase in pressure energy in the system. This is 5 to 6 orders of magnitude less energy than is required for thermal stiffness change applications \cite{Hoang2021,He2018,Park2019} and is achieved at low pressures without a vacuum, which makes it a practical option. We have also demonstrated that tunable stiffness can be used to alter the natural frequency of the system and widen the frequency range for resonant high-amplitude response. For underwater applications, we expect to see a dampening of the dynamic response and a reduction in the natural frequency due to the higher density medium. This reduction is beneficial for targeting the biological Strouhal range (0.25 to 0.4) and we plan to experimentally verify this prediction in future studies. The part of the system that changes stiffness is also fully soft, demonstrating a feasible approach for tunable soft control surfaces and propulsive fins for a new generation of soft underwater robots. 

\bibliographystyle{unsrt}
\bibliography{References}

\begin{thebibliography}{10}

\bibitem{Hoang2021}
T.~T. Hoang, J.~J.~S. Quek, M.~T. Thai, P.~T. Phan, N.~H. Lovell, and T.~N. Do.
\newblock {Soft robotic fabric gripper with gecko adhesion and variable
  stiffness}.
\newblock {\em Sensors and Actuators, A: Physical}, 323:112673, 2021.

\bibitem{Drotman2021}
D.~Drotman, S.~Jadhav, D.~Sharp, C.~Chan, and M.~T. Tolley.
\newblock {Electronics-free pneumatic circuits for controlling soft-legged
  robots}.
\newblock {\em Science Robotics}, 6(51), 2021.

\bibitem{Umedachi2016}
T.~Umedachi, V.~Vikas, and B.~A. Trimmer.
\newblock {Softworms: The design and control of non-pneumatic, 3D-printed,
  deformable robots}.
\newblock {\em Bioinspiration and Biomimetics}, 11(2), 2016.

\bibitem{Porez2014}
M.~Porez, F.~Boyer, and A.~Belkhiri.
\newblock {A hybrid dynamic model for bio-inspired soft robots - Application to
  a flapping-wing micro air vehicle}.
\newblock {\em Proceedings - IEEE International Conference on Robotics and
  Automation}, pages 3556--3563, 2014.

\bibitem{Naclerio2021}
N.~D. Naclerio, A.~Karsai, M.~Murray-Cooper, Y.~Ozkan-Aydin, E.~Aydin, D.~I.
  Goldman, and E.~W. Hawkes.
\newblock {Controlling subterranean forces enables a fast, steerable, burrowing
  soft robot}.
\newblock {\em Science Robotics}, 6(55):1--12, 2021.

\bibitem{Li2021}
G.~Li, X.~Chen, F.~Zhou, Y.~Liang, Y.~Xiao, X.~Cao, Z.~Zhang, M.~Zhang, B.~Wu,
  S.~Yin, Y.~Xu, H.~Fan, Z.~Chen, W.~Song, W.~Yang, B.~Pan, J.~Hou, W.~Zou,
  S.~He, X.~Yang, G.~Mao, Z.~Jia, H.~Zhou, T.~Li, S.~Qu, Z.~Xu, Z.~Huang,
  Y.~Luo, T.~Xie, J.~Gu, S.~Zhu, and W.~Yang.
\newblock {Self-powered soft robot in the Mariana Trench}.
\newblock {\em Nature}, 591(7848):66--71, 2021.

\bibitem{Aracri2021}
S.~Aracri, F.~Giorgio-Serchi, G.~Suaria, M.~E. Sayed, M.~P. Nemitz, S.~Mahon,
  and A.~A. Stokes.
\newblock {Soft Robots for Ocean Exploration and Offshore Operations: A
  Perspective}.
\newblock {\em Soft Robotics}, 8(6):625--639, 2021.

\bibitem{Youssef2022}
S.~M. Youssef, M.~Soliman, M.~A. Saleh, M.~A. Mousa, M.~Elsamanty, and A.~G.
  Radwan.
\newblock {Underwater Soft Robotics: A Review of Bioinspiration in Design,
  Actuation, Modeling, and Control}.
\newblock {\em Micromachines}, 13(1):1--16, 2022.

\bibitem{Sengupta2020}
D.~Sengupta, V.~Muthuram, and A.~G.~P. Kottapalli.
\newblock {Flexible Graphene-on-PDMS Sensor for Human Motion Monitoring
  Applications}.
\newblock {\em Proceedings of IEEE Sensors}, 2020-October:11--14, 2020.

\bibitem{Christianson2020}
C.~Christianson, Y.~Cui, M.~Ishida, X.~Bi, Q.~Zhu, G.~Pawlak, and M.~T. Tolley.
\newblock {Cephalopod-inspired robot capable of cyclic jet propulsion through
  shape change}.
\newblock {\em Bioinspiration and Biomimetics}, 16(1):0--9, 2020.

\bibitem{Schegg2022}
P.~Schegg and C.~Duriez.
\newblock {Review on generic methods for mechanical modeling, simulation and
  control of soft robots}.
\newblock {\em PLoS ONE}, 17(1 January 2022):1--14, 2022.

\bibitem{Arienti2013}
A.~Arienti, M.~Calisti, F.~Giorgio-serchi, and C.~Laschi.
\newblock {PoseiDRONE : design of a soft-bodied ROV with crawling , swimming
  and manipulation ability}.
\newblock {\em IEEE Xplore}, pages 7--13, 2013.

\bibitem{Al-Rubaiai2021}
M.~Al-Rubaiai.
\newblock {Enabling Soft Robotic Systems: New Solutions to Stiffness Tuning,
  Sensing, and Actuation Control}.
\newblock {\em Michigan State University}, pages 2013--2015, 2021.

\bibitem{Zhu2019}
J.~Zhu, C.~White, D.~K. Wainwright, V.~Di Santo, G.~V. Lauder, and
  H.~Bart-Smith.
\newblock Tuna robotics: A high-frequency experimental platform exploring the
  performance space of swimming fishes.
\newblock {\em Science Robotics}, 4(34):eaax4615, 2019.

\bibitem{Quinn2022}
D.~Quinn and G.~Lauder.
\newblock {Tunable stiffness in fish robotics: Mechanisms and advantages}.
\newblock {\em Bioinspiration and Biomimetics}, 17(1), 2022.

\bibitem{DEMONT1988}
M.~E. Demont and J.~M. Gosline.
\newblock { Mechanics of Jet Propulsion in the Hydromedusan Jellyfish,
  Polyorchis Pexicillatus : I. Mechanical Properties of the Locomotor Structure
  }.
\newblock {\em Journal of Experimental Biology}, 134(1):313--332, 1988.

\bibitem{Bujard}
T.~Bujard, F.~Giorgio-serchi, and G.~D. Weymouth.
\newblock {A resonant squid-inspired robot unlocks biological propulsive
  efficiency}.
\newblock {\em Science Robotics}, 2021.

\bibitem{DeLangre2008}
E.~de~Langre.
\newblock {Effects of wind on plants}.
\newblock {\em Annual Review of Fluid Mechanics}, 40:141--168, 2008.

\bibitem{Gemmell2013}
B.~J. Gemmell, J.~H. Costello, S.~P. Colin, C.~J. Stewart, J.~O. Dabiri,
  D.~Tafti, and S.~Priya.
\newblock {Passive energy recapture in jellyfish contributes to propulsive
  advantage over other metazoans}.
\newblock {\em Proceedings of the National Academy of Sciences of the United
  States of America}, 110(44):17904--17909, 2013.

\bibitem{Zhong2021}
Q.~Zhong, J.~Zhu, F.~E. Fish, S.~J. Kerr, A.~M. Downs, H.~Bart-Smith, and D.~B.
  Quinn.
\newblock {Tunable stiffness enables fast and efficient swimming in fish-like
  robots}.
\newblock {\em Science Robotics}, 6(57), 2021.

\bibitem{Huh2012}
T.~M. Huh, Y.~J. Park, and K.~J. Cho.
\newblock {Design and analysis of a stiffness adjustable structure using an
  endoskeleton}.
\newblock {\em International Journal of Precision Engineering and
  Manufacturing}, 13(7):1255--1258, 2012.

\bibitem{Li2018}
K.~Li, H.~Jiang, S.~Wang, and J.~Yu.
\newblock {A Soft Robotic Fish with Variable-stiffness Decoupled Mechanisms}.
\newblock {\em Journal of Bionic Engineering}, 15(4):599--609, 2018.

\bibitem{Behbahani2017}
S.~B. Behbahani and X.~Tan.
\newblock {Design and dynamic modeling of electrorheological fluid-based
  variable-stiffness fin for robotic fish}.
\newblock {\em Smart Materials and Structures}, 26(8):085014, 2017.

\bibitem{Leonard1960}
R.~W. Leonard, G.~W. Brooks, and H.~G. {McComb Jr}.
\newblock {Structural considerations of inflatable reentry vehicles}.
\newblock {\em {NASA Technical Note D-457}}, 1960.

\bibitem{LiMeng2021}
M.~Li, A.~Pal, A.~Aghakhani, A.~Pena-Francesch, and M.~Sitti.
\newblock {Soft actuators for real-world applications}.
\newblock {\em Nature Reviews Materials}, 0123456789, 2021.

\bibitem{Comer1963}
R.~L. Comer and S.~Levy.
\newblock {Deflections of an inflated circular-cylindrical cantilever beam}.
\newblock {\em AIAA Journal}, 1(7):1652--1655, 1963.

\bibitem{Webber1982}
J.~P.H. Webber.
\newblock {Deflections of Inflated Cylindrical Cantilever Beams Subjected To
  Bending and Torsion.}
\newblock {\em Aeronautical Journal}, 86(858):306--312, 1982.

\bibitem{Okda2019}
S.~Okda, W.~Akl, and A.~Elsabbagh.
\newblock {Structural behaviour of inflatable PVC fabric cylindrical tubes}.
\newblock {\em IOP Conference Series: Materials Science and Engineering},
  610(1):0--9, 2019.

\bibitem{Zhu2008}
Z~H Zhu, R~K Seth, and B~M Quine.
\newblock {Experimental Investigation of Inflatable Cylindrical Cantilevered
  Beams}.
\newblock {\em JP Journal of Solids and Structures}, 2(2):95--110, 2008.

\bibitem{Khayat1992}
R.~E. Khayat, A.~Derdorri, and A.~Garcia-Rjon.
\newblock {Inflation of an elastic cylindrical membrane: Non-linear deformation
  and instability}.
\newblock {\em International Journal of Solids and Structures}, 29(1):69--87,
  1992.

\bibitem{Patil2016}
A.~Patil, A.~Nordmark, and A.~Eriksson.
\newblock {Instabilities of wrinkled membranes with pressure loadings}.
\newblock {\em Journal of the Mechanics and Physics of Solids}, 94:298--315,
  2016.

\bibitem{He2018}
K.~He, S.~Liu, K.~Wang, and X.~Mei.
\newblock {Thermal characteristics of plastic film tension in roll-to-roll
  gravure printed electronics}.
\newblock {\em Applied Sciences (Switzerland)}, 8(2), 2018.

\bibitem{Park2019}
K.~B. Park, H.~T. Kim, N.~Y. Her, and J.~M. Lee.
\newblock {Variation of mechanical characteristics of polyurethane foam: Effect
  of test method}.
\newblock {\em Materials}, 12(7), 2019.

\end{thebibliography}

\end{document}